\documentclass[10pt,twocolumn,letterpaper]{article}

\usepackage{wacv}
\makeatletter
\@namedef{ver@everyshi.sty}{}
\makeatother
\usepackage{times}
\usepackage{epsfig}

\usepackage{graphicx}
\usepackage[dvipsnames]{xcolor}

\usepackage{tikz}
\usepackage{comment}
\usepackage{amsmath,amssymb} 

\usepackage{mathtools}
\usepackage{color}
\usepackage[subrefformat=parens,labelformat=parens,caption=false]{subfig}
\usepackage{graphicx}
\usepackage{booktabs}
\usepackage{multicol}
\usepackage{multirow}
\usepackage{algorithm2e}
\usepackage{xcolor}
\usepackage{enumitem}
\usepackage[normalem]{ulem}

\newcommand{\ourmodel}{FeTrIL\@\xspace}
\newcommand{\ourmodelNospace}{FeTrIL}
\newcommand{\ourmodelFc}{FeTrIL$_{fc}$\@\xspace}

\newcommand{\ourmodelone}{FeTrIL$^1$~}

\newcommand{\ourmodeloneFcNospace}{FeTrIL$^1_{fc}$}
\newcommand{\ourmodeloneFc}{FeTrIL$^1_{fc}$~}
\newcommand\scalemath[2]{\scalebox{#1}{\mbox{\ensuremath{\displaystyle #2}}}}
\definecolor{armygreen}{rgb}{0.29, 0.33, 0.13}

\def\wacvPaperID{667} 

\wacvalgorithmstrack   

\wacvfinalcopy 

\ifwacvfinal
\usepackage[breaklinks=true,bookmarks=false]{hyperref}
\else
\usepackage[pagebackref=true,breaklinks=true,colorlinks,bookmarks=false]{hyperref}
\fi

\begin{document}

\title{FeTrIL: Feature Translation for Exemplar-Free Class-Incremental Learning}

\author{
    Grégoire Petit$^{1,2}$   \\{\tt\small gregoire.petit@$\{$cea,enpc$\}$.fr} \and 
    Adrian Popescu$^{1}$     \\{\tt\small adrian.popescu@cea.fr}          \and 
    Eden Belouadah$^{1}$     \\{\tt\small hugo-schindler@orange.fr}               \and 
    David Picard$^{2}$       \\{\tt\small david.picard@enpc.fr}           \and 
    Bertrand Delezoide$^{3}$ \\{\tt\small bertrand.delezoide@amanda.com}  \and   \\
    $^1$Université Paris-Saclay, CEA, List, F-91120, Palaiseau, France\\
    $^2$LIGM, Ecole des Ponts, Univ Gustave Eiffel, CNRS, Marne-la-Vallée, France\\
    $^3$ Amanda, 34 Avenue Des Champs Elysées, F-75008, Paris, France
}

\author{
Grégoire Petit\textsuperscript{1,2}, 
Adrian Popescu\textsuperscript{1}, 
Hugo Schindler\textsuperscript{1}, 
David Picard\textsuperscript{2},
Bertrand Delezoide\textsuperscript{3}\\
 \textsuperscript{1}Université Paris-Saclay, CEA, LIST, F-91120, Palaiseau, France\\
 \textsuperscript{2}LIGM, Ecole des Ponts, Univ Gustave Eiffel, CNRS, Marne-la-Vallée, France\\
 \textsuperscript{3}Amanda, 34 Avenue Des Champs Elysées, F-75008, Paris, France\\
{\tt\small g.petit360@gmail.com, adrian.popescu@cea.fr,hugo-schindler@orange.fr}\\
{\tt\small david.picard@enpc.fr,bertrand.delezoide@amanda.com}
}

\maketitle
\thispagestyle{empty}

\begin{abstract}
Exemplar-free class-incremental learning is very challenging due to the negative effect of catastrophic forgetting.
A balance between stability and plasticity of the incremental process is needed in order to obtain good accuracy for past as well as new classes.
Existing exemplar-free class-incremental methods focus either on successive fine tuning of the model, thus favoring plasticity, or on using a feature extractor fixed after the initial incremental state, thus favoring stability.
We introduce a method which combines a fixed feature extractor and a pseudo-features generator to improve the stability-plasticity balance.
The generator uses a simple yet effective geometric translation of new class features to create representations of past classes, made of pseudo-features.
The translation of features only requires the storage of the centroid representations of past classes to produce their pseudo-features.
Actual features of new classes and pseudo-features of past classes are fed into a linear classifier which is trained incrementally to discriminate between all classes. 
The incremental process is much faster with the proposed method compared to mainstream ones which update the entire deep model.
Experiments are performed with three challenging datasets, and different incremental settings.
A comparison with ten existing methods shows that our method outperforms the others in most cases.
\ourmodel code is available at \url{https://github.com/GregoirePetit/FeTrIL}.
\end{abstract}

\begin{figure*}
\centering
    \subfloat[]{{\includegraphics[width=.24\linewidth]{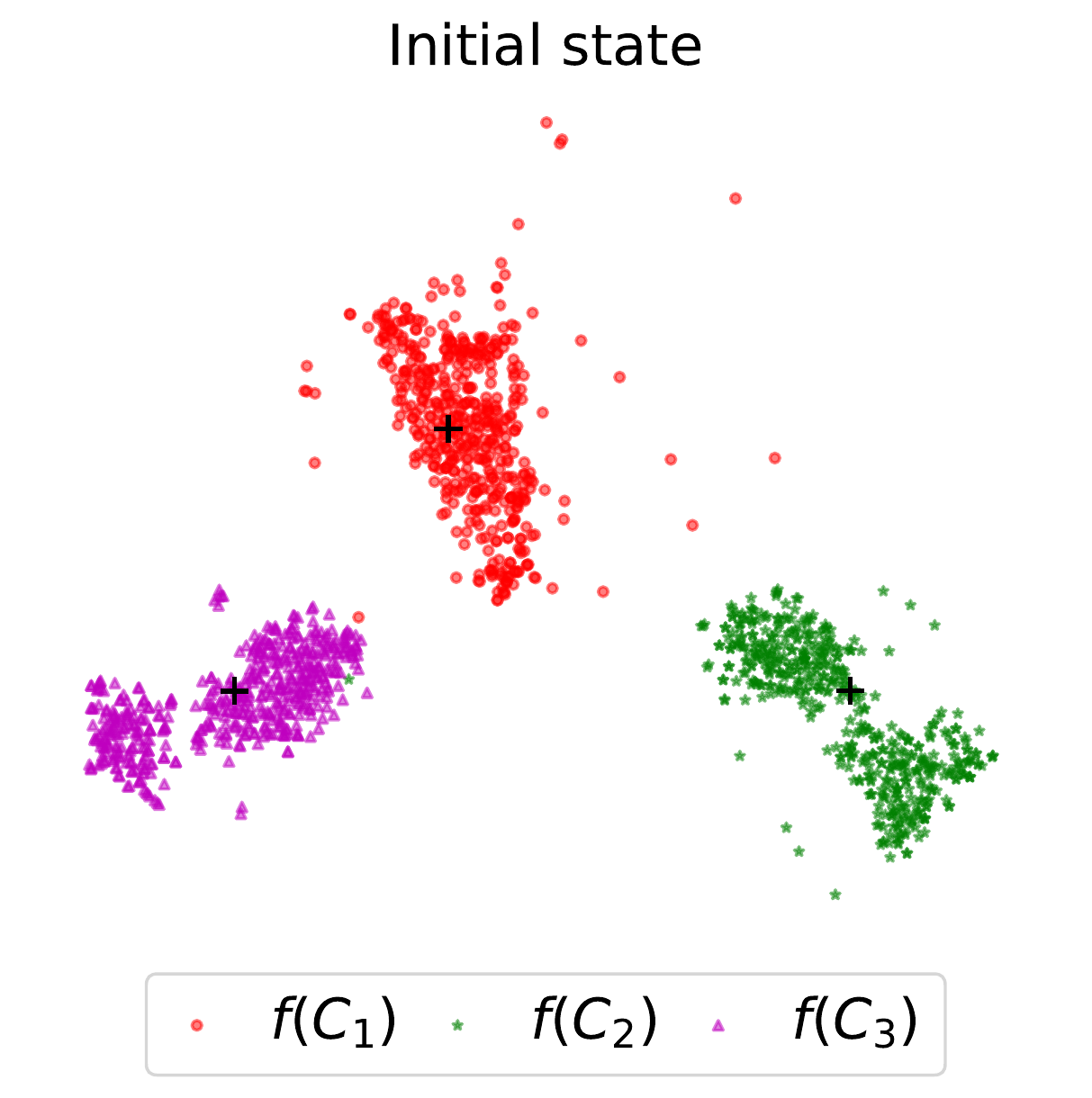} }}%
    \subfloat[]{{\includegraphics[width=.24\linewidth]{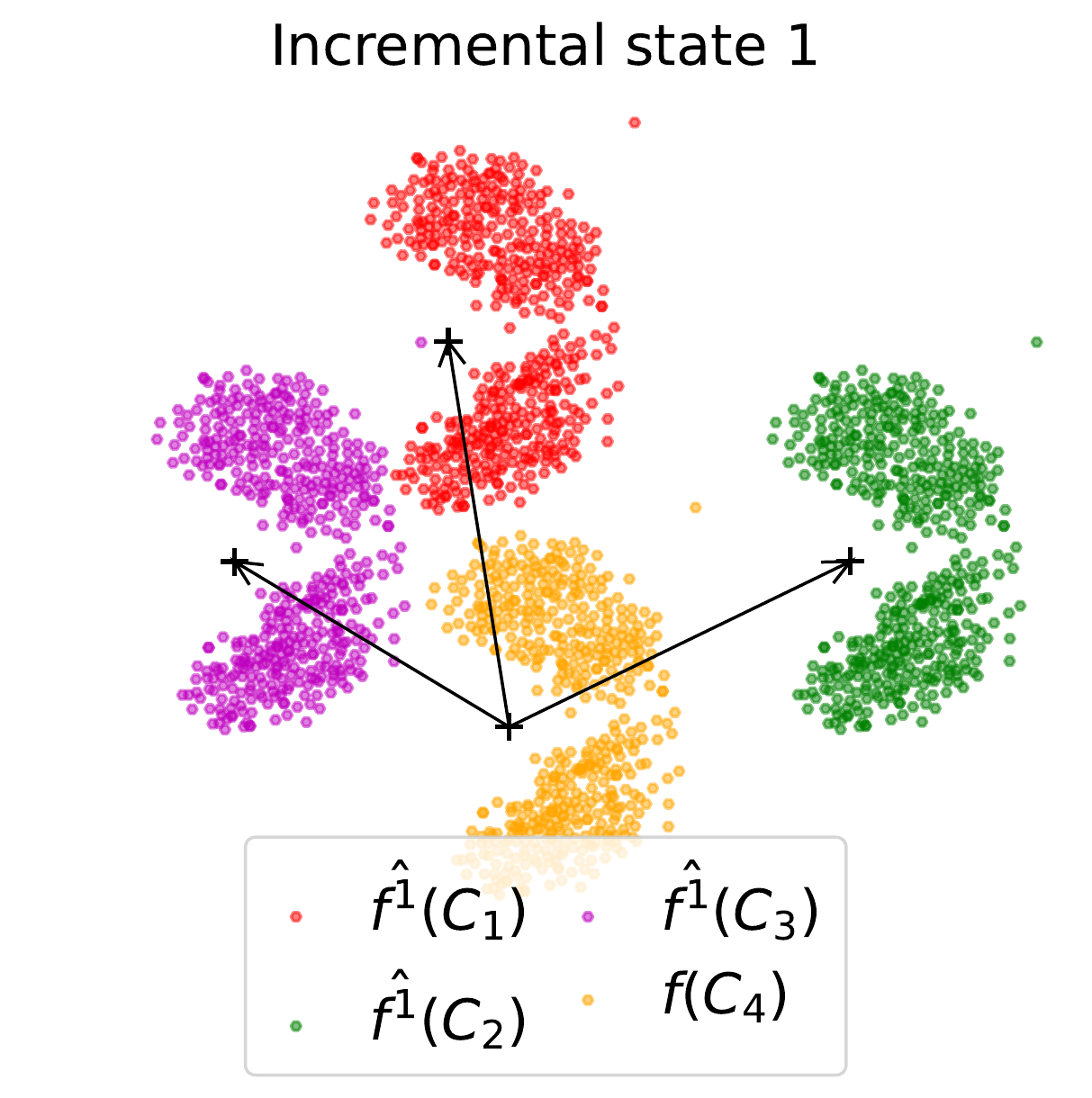} }}%
    \subfloat[]{{\includegraphics[width=.24\linewidth]{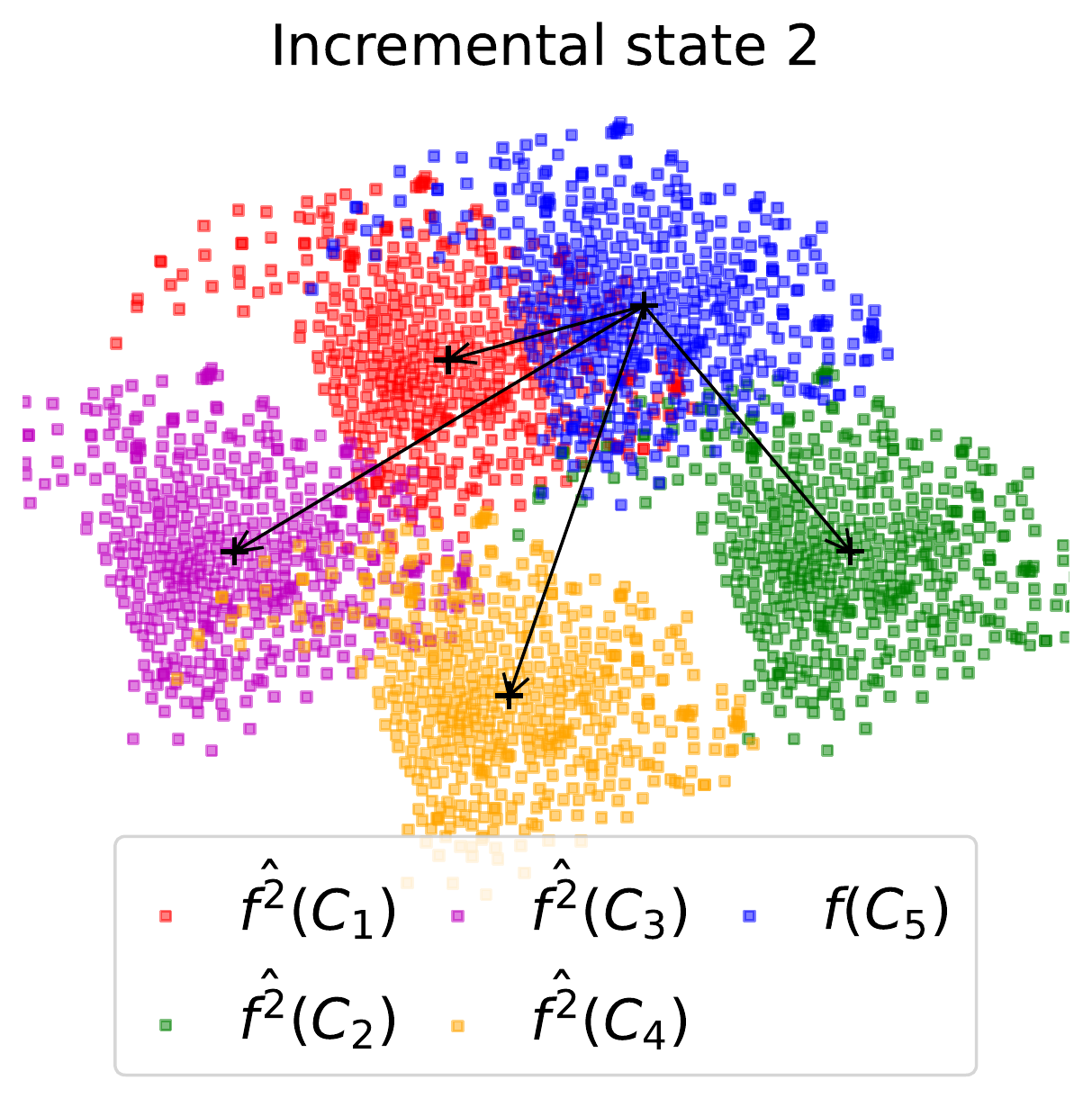} }}%
    \subfloat[]{{\includegraphics[width=.24\linewidth]{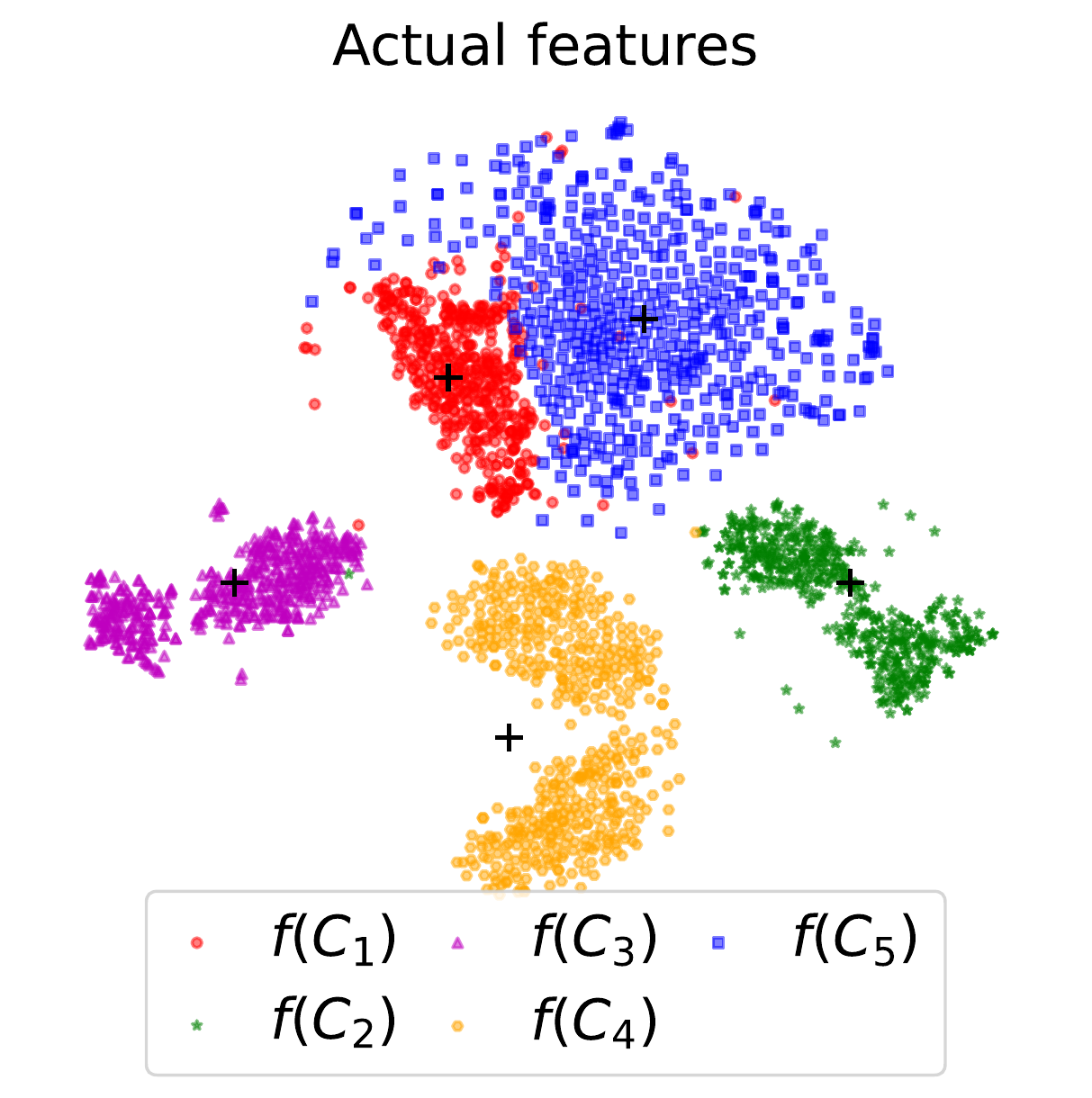} }}%

\vspace{-0cm}
	\caption{Illustration of the proposed pseudo-feature generation procedure. This toy example includes an initial state (3 classes) and two IL states (1 new class per state) in subfigures (a), (b) and (c). Subfigure (d) provides the actual features of all classes that would be available for a classical learning. The illustration uses a 2D projection of actual features. Pseudo-features of past classes are generated by geometric translation of features of the new class added in each state with the difference between the centroids of the target past class and of the new class. While imperfect, the pseudo-feature generator produces a usable representation of past classes. \textit{Best viewed in color.} \vspace{-4mm}
	}
\label{fig:teaser}
\end{figure*}


\section{Introduction}
\label{sec:introduction}
Deep learning~\cite{goodfellow2016deep} has dramatically improved the quality of automatic visual recognition, both in terms of accuracy and scale.
Current models discriminate between thousands of classes with an accuracy often close to that of human recognition, assuming that sufficient training examples are provided. 
Unlike humans, algorithms reach optimal performance only if trained with all data at once whenever new classes are learned. 
This is an important limitation because data often occur in sequences~\cite{lange2019} and their storage is costly.
Also, iterative retraining to integrate new data is computationally costly and difficult in time- or computation-constrained applications~\cite{hayes2022online,ravaglia2021tinyml}.
Incremental learning~\cite{schlimmer1986case} was introduced to reduce the memory and computational costs of machine learning algorithms. 
The main problem faced by class-incremental learning (CIL) methods is catastrophic forgetting~\cite{kemker2018measuring,mccloskey:catastrophic}, the tendency of neural nets to underfit past classes when ingesting new data.
Many recent solutions~\cite{castro2018_e2eil,hou2019_lucir,rebuffi2017_icarl,wu2019_bic,zhao2020_maintaining}, based on deep nets, use replay from a bounded memory of the past to reduce forgetting.
However, replay-based methods make a strong assumption because past data are often unavailable~\cite{venkatesan2017strategy}.
Also, the footprint of the image memory can be problematic for memory-constrained devices~\cite{ravaglia2021tinyml}. 
Exemplar-free class-incremental learning (EFCIL) methods recently gained momentum~\cite{sdc_2020,smith2021always,zhu2021class,zhu2021pass}.
Most of them use distillation~\cite{hinton2015_distillation} to preserve past knowledge, and generally favor plasticity. 
New classes are well predicted since models are learned with all new data and only a representation of past data~\cite{masana2021_study,prabhu2020gdumb,zhu2022self}.
A few EFCIL methods~\cite{belouadah2018_deesil,dhamija2021self} are inspired by transfer learning~\cite{sharif2014cnn,tan2018survey}. They learn a feature extractor in the initial state, and use it as such later to train new classifiers.
In this case, stability is favored over plasticity since the model is frozen~\cite{masana2021_study}.

We introduce \ourmodel, a new EFCIL method which combines a frozen feature extractor and a pseudo-feature generator to improve incremental performance.
New classes are represented by their image features obtained from the feature extractor.
Past classes are represented by pseudo-features which are derived from features of new classes by using a geometric translation process.
This translation moves features toward a region of the features space which is relevant for past classes.
The proposed pseudo-feature generation is adapted for EFCIL since it is simple, fast and only requires the storage of the centroids for past classes.
\ourmodel is illustrated with a toy example in Figure~\ref{fig:teaser}.
We run experiments with a standard EFCIL setting~\cite{hou2019_lucir,zhu2021class,zhu2021pass}, which consists of a larger initial state, followed by smaller states which include the same number of classes.  
Results show that the proposed approach has better behavior compared to ten existing methods, including very recent ones.

\section{Related Work}
\label{sec:related}
CIL algorithms are needed when data arrive sequentially and/or computational constraints are important~\cite{hayes2022online,lange2019,masana2021_study,parisi2019_continual}. 
Their objective is to ensure a good balance between plasticity, i.e. integration of new information, and stability, i.e. preservation of knowledge about past classes~\cite{mermillod2013_stability_plasticity}.
This is challenging because the lack of past data leads to catastrophic forgetting, i.e. the tendency of neural networks to focus on newly learned data at the expense of past knowledge~\cite{mccloskey:catastrophic}.
Recent reviews of CIL~\cite{belouadah2021_study,masana2021_study} show that a majority of methods replay samples of past classes to mitigate  forgetting~\cite{castro2018_e2eil,hou2019_lucir,rebuffi2017_icarl,zhao2020_maintaining}.
One advantage here is that the network architecture remains constant throughout the incremental process. 
However, these methods have two major drawbacks: (1) 
First, the assumption that past samples are available is strong since in many cases past data cannot be stored due, for instance, to privacy restrictions~\cite{venkatesan2017strategy} and
(2) the memory footprint of the stored images is high. 

Here, we investigate exemplar-free CIL, with focus on methods which keep the network size constant.
This setting is very challenging since it imposes strong constraints on both memory and computational costs. 
A majority of existing methods use regularization to update the deep model for each incremental step~\cite{masana2021_study}, and adapt  distillation~\cite{hinton2015_distillation} to preserve past knowledge by penalizing variations for past classes during model updates.
Note that, while some of the distillation-based methods were introduced in an exemplar-based CIL (EBCIL) setting, many of them are also applicable to EFCIL.
This approach to CIL was popularized by iCaRL~\cite{rebuffi2017_icarl}, itself inspired by learning without forgetting (LwF)~\cite{li2016_lwf}.
Distillation was later refined and complemented with other components to improve the plasticity-stability compromise. 
LUCIR~\cite{hou2019_lucir} applies distillation on features instead of raw classification scores to preserve the geometry of past classes, and an inter-class separation to maximize the distances between past and new classes. 
The problem was partially addressed by adding specific class separability components in ~\cite{douillard2020podnet,hou2019_lucir}.
Distillation-based methods need to store the current and the preceding model for incremental updates.
Their memory footprint is larger compared to methods which do not use distillation~\cite{masana2021_study}.

Another important problem in CIL is the semantic drift between incremental states.
Auxiliary classifiers were introduced in~\cite{liu2020more} to reduce the effect of forgetting.
ABD~\cite{smith2021always} uses image inversion to produce pseudo-samples of past classes.
The method is interesting but image inversion is difficult for complex datasets.
Another interesting solution is proposed in~\cite{sdc_2020}, where the features drift between incremental steps is estimated from that of new classes.
Recent EFCIL approaches~\cite{zhu2021class,zhu2021pass,zhu2022self} use past class prototypes in conjunction with distillation to improve performance. 
Prototype augmentation is proposed in PASS~\cite{zhu2021pass} to improve the discrimination of classes learned in different incremental states.  
Feature generation for past classes is introduced in IL2A~\cite{zhu2021class} by leveraging information about the class distribution.
This approach is difficult to scale-up because a covariance matrix needs to be stored for each class.
A prototype selection mechanism is introduced in SSRE~\cite{zhu2022self} to better discriminate past from new classes. 
\ourmodel shares the idea of using class prototypes with~\cite{sdc_2020,zhu2021class,zhu2021pass,zhu2022self}.
An important difference is that we freeze the model after the initial state, while the other methods deploy more sophisticated mechanisms to integrate prototypes in a knowledge distillation process.
Past comparative studies~\cite{belouadah2021_study,masana2021_study} found that, while appealing in theory, distillation-based methods underperform in EFCIL, particularly for large-scale datasets. 
Second, since the representation space is fixed, a simple geometric translation of actual features of new classes is sufficient to produce usable pseudo-features.
In contrast, IL2A~\cite{zhu2021class}, the work which is closest to ours, needs to store a covariance matrix per class to obtain optimal performance. 
Third, the use of a fixed extractor simplifies the training process since only the final linear layer is trained, compared to a fine tuning of the backbone model required by recent methods which use prototypes and feature generation. 

Another line of work takes inspiration from transfer learning~\cite{neyshabur2020being,sharif2014cnn} to tackle EFCIL.
A feature extractor is trained in the initial non-incremental state and fixed afterwards. 
Then, an external classification layer is updated in each incremental state to integrate new classes. 
The nearest class mean (NCM)~\cite{mensink2013distance} was used in ~\cite{rebuffi2017_icarl}, linear SVMs~\cite{fabian2012_scikitlearn} were used in~\cite{belouadah2018_deesil} and extreme value machines~\cite{rudd2017extreme} were recently tested by~\cite{dhamija2021self}. 
The advantages of transfer-learning methods are their simplicity, since only the classification layer is updated, and their lower memory requirement, since they need a single deep model to function.
These methods give competitive performance compared to distillation-based ones in EFCIL, particularly at scale~\cite{belouadah2021_study}.
However, features are not updated, and they are sensitive to large domain shifts between incremental tasks~\cite{lange2019}. 
Equally, existing transfer-learning inspired works do not sufficiently address inter-class separability, which is in focus here.

Class prototypes creation was studied in other learning settings than CIL.
A very interesting method focused on few-shot learning was proposed in~\cite{das2019two}. 
A distance-based classifier which uses an approximation of the Mahalanobis distance is proposed.
The means and variances of new classes are predicted using two supplementary neural networks. 
While adapted for few-shot learning, such an approach is not fully adapted in CIL.
First, the supplementary neural networks require a large number of supplementary parameters. 
This is a disadvantage here, since CIL methods are needed in computationally-constrained environments. 
Second, we do not focus on few-shot learning and the means of past classes are well-placed in the representation space. 

\section{Proposed Method}
\label{sec:method}

\begin{figure*}[t]
	\centering
	\includegraphics[width=0.75\linewidth, trim={0cm 0cm 0cm 0cm}]{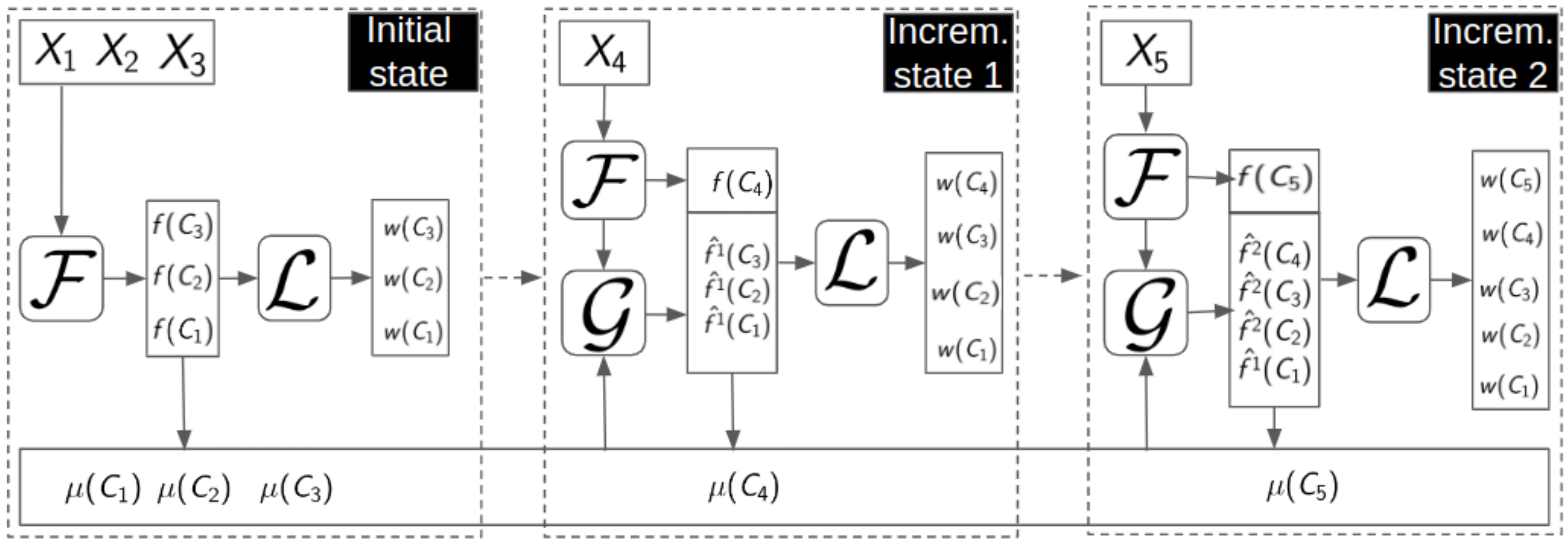} 
	\caption{\ourmodel overview for a toy example with an initial state (3 classes) and two incremental states (1 class per state). The feature extractor $\mathcal{F}$ is trained in the initial state, using sets of data $X_1, X_2, X_3$, and then frozen afterwards. The generator $\mathcal{G}$ uses features $f(C_n)$ of the new class extracted with $\mathcal{F}$ and prototypes of past classes $\mu(C_p)$ to generate pseudo-features of past classes $\hat{f^t}(C_p)$ in the $t^{th}$ state. Prototypes ($\mu(C_i)$) are the centroids of all classes (past and new). They are learned when classes are first seen and then stored throughout the IL process. A linear classifier $\mathcal{L}$ is used to learn classification weights $w(C_i)$ for all seen classes (past and new).   
	}
	\vspace{-4mm}
	\label{fig:overview}
\end{figure*}

The objective of CIL is to learn a total of $N$ classes which appear sequentially during training. 
This process includes an initial state (0) and $T$ incremental ones. 
New classes need to be recognized alongside past classes which were learned in previous states.
We focus on the exemplar-free CIL setting~\cite{rebuffi2017_icarl,smith2021always,sdc_2020,zhu2022self}, which assumes that no past images can be stored.
This scenario is more challenging than exemplar-based CIL since catastrophic forgetting needs to be tackled without resorting to replay~\cite{masana2021_study}. 
There is no intersection between the classes learned in different incremental states.
Unlike task IL~\cite{van2019three}, the boundaries between different states are not known at test time. 

The global functioning of \ourmodel is illustrated in Figure~\ref{fig:overview}. 
It uses a feature extractor, a pseudo-feature generator based on geometric translation, and an external classification layer in order to address EFCIL.
Inspired by transfer-learning based CIL~\cite{belouadah2018_deesil,rebuffi2017_icarl}, the feature extractor $\mathcal{F}$ is frozen after the initial state.
This ensures a stable representation space through the entire CIL process. 
Given that images of past classes cannot be stored in EFCIL, a generator $\mathcal{G}$ is used to produce pseudo-features of past classes ($\hat{f^t}(C_p)$).
$\mathcal{G}$ takes features of new classes ($f(C_n)$) and prototypes of past and new classes ($\mu(C_p)$, $\mu(C_n)$) as inputs. 
A linear classifier $L$ combines features and pseudo-features to jointly train classifiers for all seen classes (past and new). 
The pseudo-features generation is crucial since it enables class discrimination across all incremental states.
The hypotheses made here are that: 
(1) while imperfect, the pseudo-features still produce effective representations of past classes,
and
(2) using a frozen extractor in combination with a generator in EFCIL is preferable to mainstream distillation-based methods~\cite{sdc_2020,zhu2021class,zhu2021pass,zhu2022self}.
These hypotheses are tested through the extensive experiments in Section~\ref{sec:evaluation}.
We present the main components of \ourmodel in the next subsections. 

\subsection{Generation of pseudo-features}
\label{subsec:gen}
The pseudo-feature generator, illustrated in Figure~\ref{fig:teaser}, produces effective representations of past classes.
Existing approaches which generate past data rely on methods such as generative adversarial networks~\cite{he2018_generative}, image inversion~\cite{smith2021always}, or covariance-based past class models~\cite{zhu2021class}.
We propose a much simpler alternative which is defined as: \vspace{-2mm}

\begin{equation}
    {\hat{f^t}}(c_p) = f(c_n) + \mu(C_p) - \mu(C_n)
\label{eq:generator}
\end{equation}
with: 
$C_p$ - target past class for which pseudo-features are needed; 
$C_n$ - new class for which images $b$ are available;
$f(c_n)$ - features of a sample $c_n$ of class $C_n$ extracted with $\mathcal{F}$;
$\mu(C_p), \mu(C_n)$ - mean features of classes $C_p$ and $C_n$ extracted with $\mathcal{F}$;
${\hat{f^t}}(c_p)$ - pseudo-feature vector of a pseudo-sample $c_p$ of class $C_p$ produced in the $t^{th}$ incremental state.

Eq.~\ref{eq:generator} translates the value of each dimension with the difference between the values of the corresponding dimension of $\mu(C_p)$ and $\mu(C_n)$. 
It creates a pseudo-feature vector situated in the region of the representation space associated to target class $C_p$ based on actual features of a new class $f(C_n)$.
The computational cost of generation is very small since it only involves additions and subtractions. 
$\mu(C_p)$ is needed to drive the geometric translation toward a region of the representation space which is relevant for $C_p$.
Centroids are computed when classes occur for the first time and then stored.
Their reuse is possible because $\mathcal{F}$ is fixed after the initial step and its associated features do not evolve.

\subsection{Selection of pseudo-features}
\label{subsec:selection}
Eq.~\ref{eq:generator} translates the features for a single sample. 
If each class is represented by $s$ samples, the generation process needs to be repeated $s$ times.
The overview of \ourmodel (Figure~\ref{fig:overview}) and of the pseudo-feature generation (Figure~\ref{fig:teaser}) use a minimal example which adds a single class per IL state.
When CIL states include several classes $ C_n$, the $s$ pseudo-features of each class $C_p$ can be obtained using different strategies, depending on how features of new classes are used. 
We deploy the following strategies:

\begin{itemize}[nosep,leftmargin=*]
    \item \ourmodelNospace$^k$: $s$ features are transferred from the $k^{th}$ similar new class of each past class $C_p$. Similarities between the target $C_p$ and the $C_n$ available in the current state is computed using the cosine similarity between the centroids of each pair of classes.  Experiments are run with different values of $k$ to assess if a variable class similarity has a significant effect on EFCIL performance. Since translation is based on a single new class, the distribution of pseudo-features will be similar to that of features of $C_n$, but in the region of the representation space around $\mu(C_p)$. 
    \item \ourmodelNospace$^{rand}$: $s$ features are randomly selected from all new classes. This strategy assesses whether a more diversified source of features from different $C_n$ produces an effective representation of class $C_p$. 
    \item \ourmodelNospace$^{herd}$: $s$ features are selected from any new class based on a herding algorithm~\cite{max2009_herding}. It assumes that sampling should include features which produce a good approximation of the past class. Herding was introduced in exemplar-based CIL in order to obtain an accurate approximation of each class by using only a few samples~\cite{rebuffi2017_icarl} and its usefulness was later confirmed \cite{belouadah2021_study,hou2019_lucir,wu2019_bic}. It is adapted here to obtain a good approximation of the sample distribution of $C_p$ with $s$ pseudo-features.
\end{itemize}

The comparison of these different strategies will allow us to determine whether the geometric translation of features is prevalent, or if a particular configuration of the features around the centroid of the target past class is needed. 

\subsection{Linear classification layer training}
\label{subsec:layer}
We assume that the CIL process is in the $t^{th}$ CIL state, which includes $P$ past classes and $N$ new classes.
The combination of the feature generator (Subsection~\ref{subsec:gen}) and selection (Subsection~\ref{subsec:selection}) provides a set ${\hat{f^t}}(C_p)$ of $s$ pseudo-features for each class $C_p$. 
The objective is to train a linear classifier for all $P+N$ seen classes which takes pseudo features of past classes and actual features of new classes as inputs.
This linear layer is defined as:
\begin{equation}
\scalemath{0.9}{
    \mathcal{W}^t = \{w^t(C_1),..., w^t(C_P), w^t(C_{P+1}),..., w^t(C_{P+N})\}
}
    \label{eq:linear}
\end{equation}
with: $w^t$ - the weight of known classes in the $t^{th}$ CIL state. 

$\mathcal{W}^t$ can be implemented using different classifiers, and we instantiate two versions in Section~\ref{sec:evaluation}: 
(1) \ourmodel using LinearSVCs~\cite{fabian2012_scikitlearn} as external classifiers, 
and 
(2) \ourmodelFc using a fully-connected layer to enable end-to-end training. 

\section{Evaluation}
\label{sec:evaluation}
We evaluate \ourmodel by using a comprehensive EFCIL evaluation scenario~\cite{zhu2021class,zhu2021pass,zhu2022self}. 
This setting includes four datasets and CIL states of different size.

\textbf{Datasets.} We use four public datasets:
(1) CIFAR-100~\cite{krizhevsky2009_cifar100} - 100 classes, 32x32 pixels images, 500 and 100 images/class for training and test;
(2) TinyImageNet~\cite{le2015tiny} - 200 leaf clases from ImageNet, 64x64 pixels images, 500 and 50 for training and test;
(3) ImageNet-Subset - 100 classes subset of ImageNet LSVRC dataset~\cite{olga2015_ilsvrc}, ~1300 and 50 for training and test;
(4) ILSVRC - full dataset from~\cite{olga2015_ilsvrc}.

\textbf{Incremental setting.} We use a classical EFCIL protocol from~\cite{zhu2021class,zhu2021pass,zhu2022self}. 
The number of classes in the initial state is larger, and the rest of the classes are evenly distributed between incremental states. 
CIFAR-100 and ImageNet-Subset are tested with: (1) 50 initial classes and 5 IL states of 10 classes, (2) 50 initial classes and 10 IL states of 5 classes, (3) 40 initial classes and 20 states of 3 classes, and (4) 40 initial classes and 60 states of 1 class. 
Compared to~\cite{zhu2021class,zhu2021pass,zhu2022self}, configurations (1) and (3) for ImageNet-Subset are added for more consistent evaluation.
TinyImageNet is tested with 100 initial classes and the other classes distributed as follows: (1) 5 states of 20 classes, (2) 10 states of 10 classes, (3) 20 states of 5 classes, and (4) 100 states of 1 class.
Configuration (4) is interesting since it enables one class increments.
It cannot be deployed for any of the compared EFCIL methods since they require at least two classes per increment to update models. 
ILSVRC is tested with 500 initial classes, and the other 500 split evenly among $T\in \{5,10,20\}$ states.
This enables a comprehensive comparison of the methods in varied EFCIL configurations.
Naturally, task IDs are not available at test time. 

\textbf{Compared methods.} We use the following EFCIL methods in evaluation: EWC~\cite{kirkpatrick2017overcoming}, LwF-MC~\cite{rebuffi2017_icarl}, DeeSIL~\cite{belouadah2018_deesil}, LUCIR~\cite{hou2019_lucir}, MUC~\cite{liu2020more}, SDC~\cite{sdc_2020}, PASS~\cite{zhu2021pass}, ABD~\cite{smith2021always}, IL2A~\cite{zhu2021class}, SSRE~\cite{zhu2022self}.
As we discussed in Section~\ref{sec:related}, these methods cover a large variety of EFCIL approaches.
The inclusion of recent works~\cite{zhu2021class,zhu2021pass,zhu2022self} is important to situate our contribution with respect to current EFCIL trends. 
While focus is on EFCIL, we follow~\cite{zhu2022self} and include a comparisonwith EBCIL methods. 
We test our method against the recent AANets approach~\cite{liu2021aanets}, and against the EBCIL methods to which AANETS was added (LUCIR~\cite{hou2019_lucir}, Mnemonics~\cite{mnemonics_2020}, PODNet~\cite{douillard2020podnet}).
Whenever available, results of compared methods marked with $^*$ are reproduced either from their initial paper or from~\cite{zhu2022self} for EFCIL or from~\cite{liu2021aanets} for EBCIL. 
The other results are recomputed using the original 
configurations of the methods.

\textbf{Implementation details.} Following~\cite{rebuffi2017_icarl,zhu2021class,zhu2021pass,zhu2022self}, we use ResNet-18~\cite{he2016_resnet} in all experiments.
\ourmodel initial training is done uniquely with images of initial classes to ensure comparability with existing methods.
The feature extractor is trained in the initial state and then frozen for the reminder of the IL process. 
We implement a supervised training with cross-entropy loss, SGD optimization, a batch size of 128, for a total of 160 epochs.
The initial learning rate is 0.1, and it is decayed by 0.1 after every 50 epochs. 
To ensure comparability, classes are assigned to IL states using the same random seed as in the compared methods~\cite{hou2019_lucir,zhu2021pass,zhu2021class,zhu2022self}. 

We provide implementation details for the final layer (Eq.~\ref{eq:linear}) introduced in Subsection~\ref{subsec:layer}.
The hyperparameters of the classification layers were optimized on a pool of 50 classes selected randomly from ImageNet, but disjoint from ILSVRC or ImageNet-Subset. 
L2-normalization is applied before the linear layer.
The LinearSVC layer included in \ourmodelone uses 1.0 and 0.0001 for 
regularization and the tolerance parameters.
The number of samples is higher than the dimensionality of the features, and we solve the primal rather than the dual optimization problem.
The classifiers are then trained using a standard one against the rest procedure.
In Subsection~\ref{subsec:analysis}, we also test a one-vs-many strategy to accelerate incremental updates.
The second variant, \ourmodeloneFcNospace, using a fully-connected layer as final layer, and implements an end-to-end training strategy. 
\ourmodeloneFcNospace is trained for 50 epochs with an initial learning rate of 0.1, 0.1 decay, and 10 epochs patience.

\textbf{Evaluation metric.} The average incremental accuracy, widely used in CIL ~\cite{masana2021_study,rebuffi2017_icarl}, is the main evaluation measure.
For comparability with~\cite{zhu2021class,zhu2021pass,zhu2022self}, it is computed as the average accuracy of all states, including the initial one. 
We equally provide per-state accuracy curves to have a more detailed view of the accuracy evolution during the CIL process. 
Following~\cite{zhu2022self}, we run each configuration of \ourmodel three times and report the averaged results.

\subsection{Results}
\label{subsec:results}

\begin{table*}[t]
\begin{center}
\resizebox{0.95\textwidth}{!}{
\begin{tabular}{@{\kern0.5em}llccccccccccccccccccc@{\kern0.5em}}
        \toprule
        \multirow{2}{*}{\textbf{CIL Method}}
            & \multicolumn{4}{c}{CIFAR-100}
            && \multicolumn{4}{c}{TinyImageNet}
            && \multicolumn{4}{c}{ImageNet-Subset}
            && \multicolumn{3}{c}{ImageNet}
        \\ \cmidrule(lr){2-5} \cmidrule(lr){7-10} \cmidrule(l){12-15} \cmidrule{17-19}  
        
            & \multicolumn{1}{c}{\textit{T}=5}
            & \multicolumn{1}{c}{\textit{T}=10}
            & \multicolumn{1}{c}{\textit{T}=20}
            & \multicolumn{1}{c}{\textit{T}=60}
            &
            & \multicolumn{1}{c}{\textit{T}=5}
            & \multicolumn{1}{c}{\textit{T}=10}
            & \multicolumn{1}{c}{\textit{T}=20}
            & \multicolumn{1}{c}{\textit{T}=100}
            &
            & \multicolumn{1}{c}{\textit{T}=5}
            & \multicolumn{1}{c}{\textit{T}=10}
            & \multicolumn{1}{c}{\textit{T}=20}
            & \multicolumn{1}{c}{\textit{T}=60}
            &
            & \multicolumn{1}{c}{\textit{T}=5}
            & \multicolumn{1}{c}{\textit{T}=10}
            & \multicolumn{1}{c}{\textit{T}=20}
            
        \\ \midrule

EWC$^*$~\cite{kirkpatrick2017overcoming} \small (PNAS'17)    & 24.5 & 21.2 & 15.9 & x && 18.8 & 15.8 & 12.4 & x && -    & 20.4 & - & x  && - & - & - \\
LwF-MC$^*$~\cite{rebuffi2017_icarl} \small (CVPR'17) & 45.9 & 27.4 & 20.1 & x && 29.1 & 23.1 & 17.4 & x && -    & 31.2 & - & x  && - & - & - \\    
LUCIR \small (CVPR'19)      & 51.2 & 41.1 & 25.2 & x && 41.7 & 28.1 & 18.9 & x && 56.8 & 41.4 & 28.5  & x && 47.4 & 37.2 & 26.6 \\  
MUC$^*$~\cite{liu2020more} \small (ECCV'20)    & 49.4 & 30.2 & 21.3 & x && 32.6 & 26.6 & 21.9 & x && -    & 35.1 & - & x && - & - & - \\   
SDC$^*$~\cite{sdc_2020} \small (CVPR'20)    & 56.8 & 57.0 & 58.9 & x &&   -  & -    &-  & x  && -    & 61.2 & - & x && - & - & - \\   
ABD$^*$~\cite{smith2021always} \small (ICCV'21)    & 63.8 & 62.5 & 57.4 & x &&   -  & -    &- & x    && -    & - & - & x && - & - & - \\ 
PASS$^*$~\cite{zhu2021pass} \small (CVPR'21)   & 63.5 & 61.8 & 58.1 & x && 49.6 & 47.3 & 42.1 & x && 64.4   & 61.8 & {51.3} & x && - & - & - \\ 
IL2A$^*$~\cite{zhu2021class} \small (NeurIPS'21)& \underline{66.0} & 60.3 & 57.9 & x && 47.3 & 44.7 & 40.0  & x && - & - & - & x  && - & - & - \\ 
SSRE$^*$~\cite{zhu2022self} \small (CVPR'22)   & 65.9 & \underline{65.0} & \textbf{61.7} & x && {50.4} & {48.9} & {48.2} & x && -    & {67.7} &  - & x && - & - & - \\ 
\hline
DeeSIL~\cite{belouadah2018_deesil} \small (ECCVW'18)    & 60.0 & 50.6 & 38.1 & x && 49.8 & 43.9 & 34.1 & x &&  {67.9} & 60.1 & 50.5 & x && 61.9 & 54.6 & 45.8 \\

DSLDA~\cite{hayes2020_deepslda} \small (CVPRW'20)   & 64.0 & 63.8 & 60.8 & \textbf{60.5} && \underline{53.1} & \underline{52.9} & \textbf{52.8} & \textbf{52.6} && \underline{71.3}  & \textbf{71.2} &  \textbf{71.0} & \textbf{70.8} && \underline{64.0} & \underline{63.8} & \underline{63.6} \\

\ourmodelone  & \textbf{66.3} & \textbf{65.2} & \underline{61.5} & \underline{59.8} && \textbf{54.8} & \textbf{53.1} & \underline{52.2} & \underline{50.2} && \textbf{72.2} & \textbf{71.2} & \underline{67.1} & \underline{65.4} && \textbf{66.1} & \textbf{65.0} & \textbf{63.8} \\     
\hline
\end{tabular}
}
\end{center}
\vspace{-2mm}
	\caption{Average top-1 incremental accuracy in EFCIL with different numbers of incremental steps.
	\ourmodelone results are reported with pseudo-features translated from the most similar new class.
	"-" cells indicate that results were not available (see supp. material for details). "x" cells indicate that the configuration is impossible for that method. 
	\textbf{Best results - in bold}, \underline{second best - underlined}.\vspace{-4.3mm}}
\label{tab:main_results}
\end{table*}

\textbf{Comparison to existing EFCIL methods.} The results from Table~\ref{tab:main_results} show that \ourmodelone outperforms all compared methods in 11 tested configurations out of 12. 
It is also close to the best in the remaining one.
The second best results are obtained with the very recent SSRE method~\cite{zhu2022self}.
\ourmodelone and SSRE accuracies are close to each other for CIFAR-100, with relative differences between 0.4 and -0.2.
The performance gain brought by \ourmodel is of over 4 and 3 top-1 accuracy points for TinyImageNet and ImageNet-Subset, respectively. 
PASS~\cite{zhu2021pass} and IL2A~\cite{zhu2021class}, two other recent EFCIL methods, have lower average performance.
We note that EFCIL performance boost was recently reported, with methods such as PASS, IL2A, SSRE.
These methods combine knowledge distillation and sophisticated mechanisms for dealing with the stability-plasticity dilemma.
In contrast, our method uses a fixed feature extractor and a lightweight pseudo-feature generator. 
\ourmodel only optimizes a linear classification layer, while compared recent methods use backpropagation of the entire model, and need much more computational resources and time to perform the IL process.
A more in-depth discussion of complexity is proposed in Subsection~\ref{subsec:analysis}.
Performance of the ILSVRC dataset is also very interesting. 
Direct comparison to PASS or SSRE is impossible since these methods were not tested at scale.
However, we can safely assume that \ourmodelone is better given PASS and SSRE accuracy for the simpler ImageNet-Subset. 
ILSVRC results show that the simple method proposed here is effective for a high range of classes. 
Interestingly, ILSVRC performance is stabler compared to smaller datasets since the pool of new classes available for pseudo-features generation is larger.

\textbf{Comparison to a transfer-learning baseline.} DeeSIL~\cite{belouadah2018_deesil} is a simple application of transfer learning to EFCIL.
It has no class separability mechanism across different incremental states since classifiers are learned within each state. 
The need for global separability, included in \ourmodel, is shown by the comparison of short and long CIL processes.
DeeSIL~\cite{belouadah2018_deesil} performance is good for $T=5$ because each class is trained against enough other classes, but drops significantly for $T=20$, when there are few new classes.
The important performance gain brought by \ourmodel highlights the importance of class separability. 

\textbf{Behavior for minimal incremental updates.}
Compared EFCIL methods can only be updated with a minimum of two classes per CIL state since they use discriminative classifiers, which require both positive and negative samples.
In practice, it is interesting to enable updates once each new class is available.
This is possible with \ourmodel because pseudo-features can all originate from a single new class.
Results in the right columns of CIFAR-100, TinyImageNet and ImageNet-Subset from Table~\ref{tab:main_results} show that the accuracy obtained in with one class increments is close to that observed for $T=20$. 
This highlights the robustness of \ourmodel with respect to frequent updates. 


\textbf{Influence of the final classification layer.} 
\ourmodelone compares favorably with \ourmodeloneFcNospace.
LinearSVC gives better performance than a fully-connected layer, particularly for a large number of incremental steps. 
However, \ourmodeloneFc is also competitive, and outperforms existing methods in a majority of configurations.

\begin{figure*}
\centering
    {{\includegraphics[width=.32\linewidth]{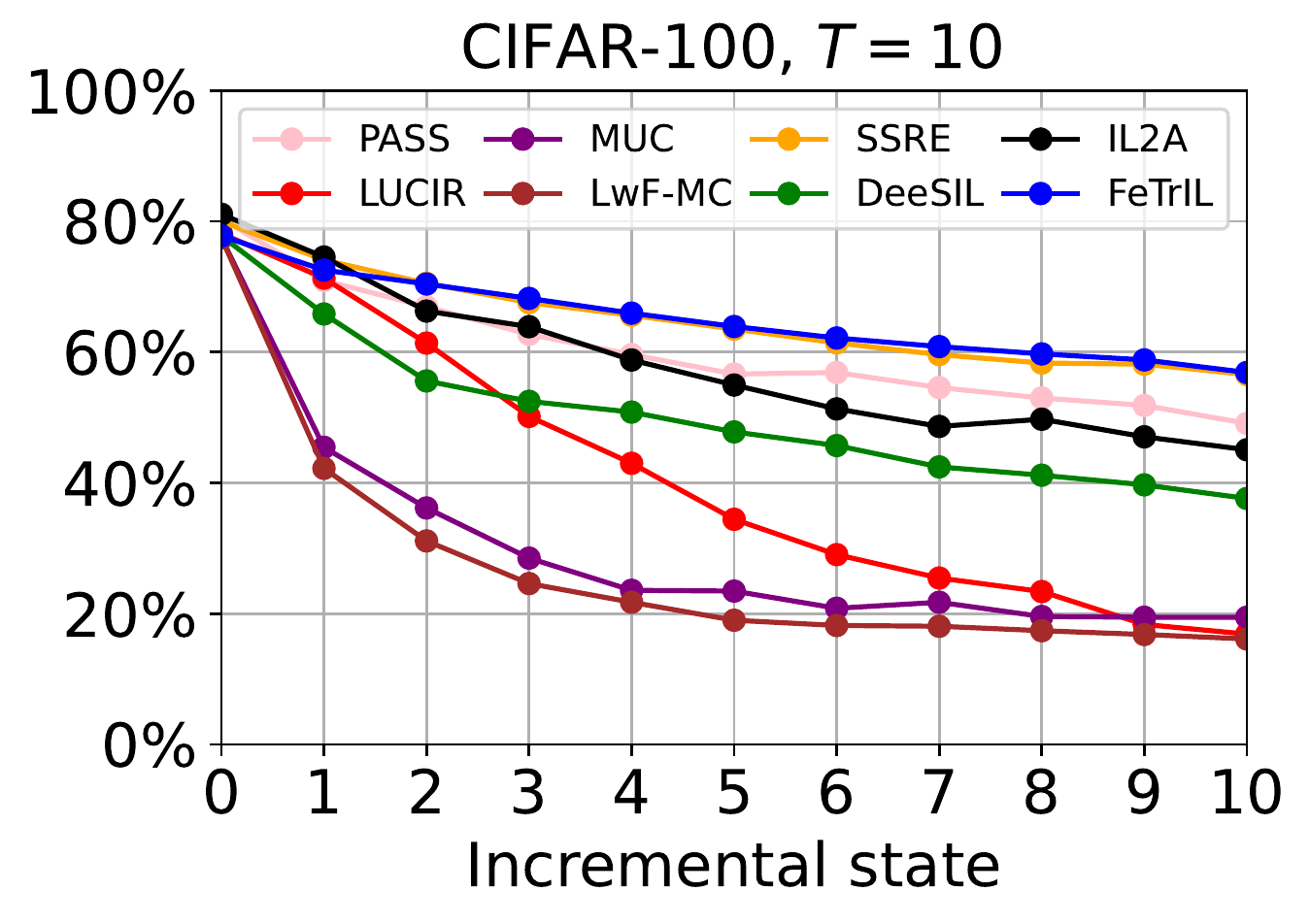} }}%
    {{\includegraphics[width=.32\linewidth]{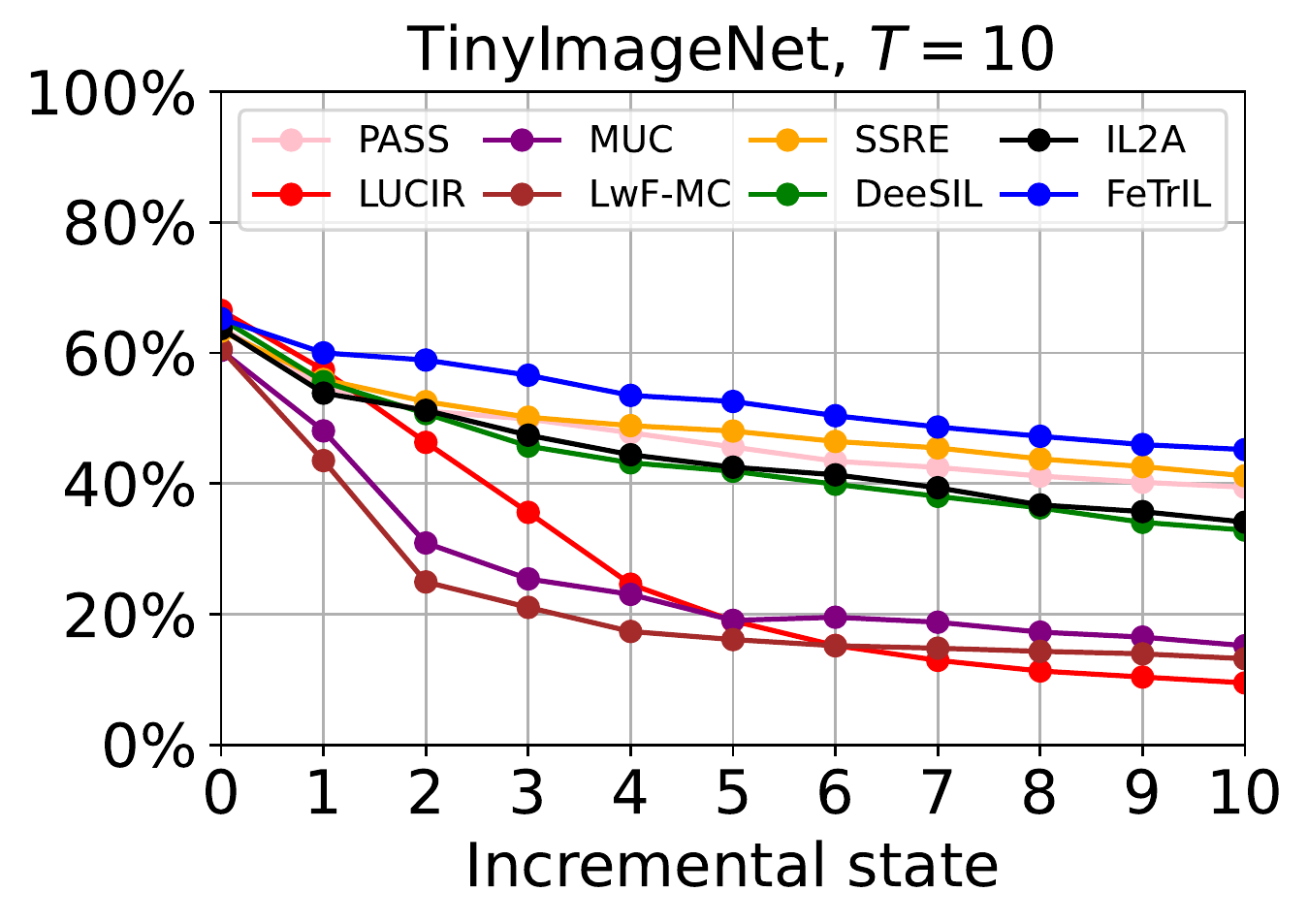} }}%
    {{\includegraphics[width=.32\linewidth]{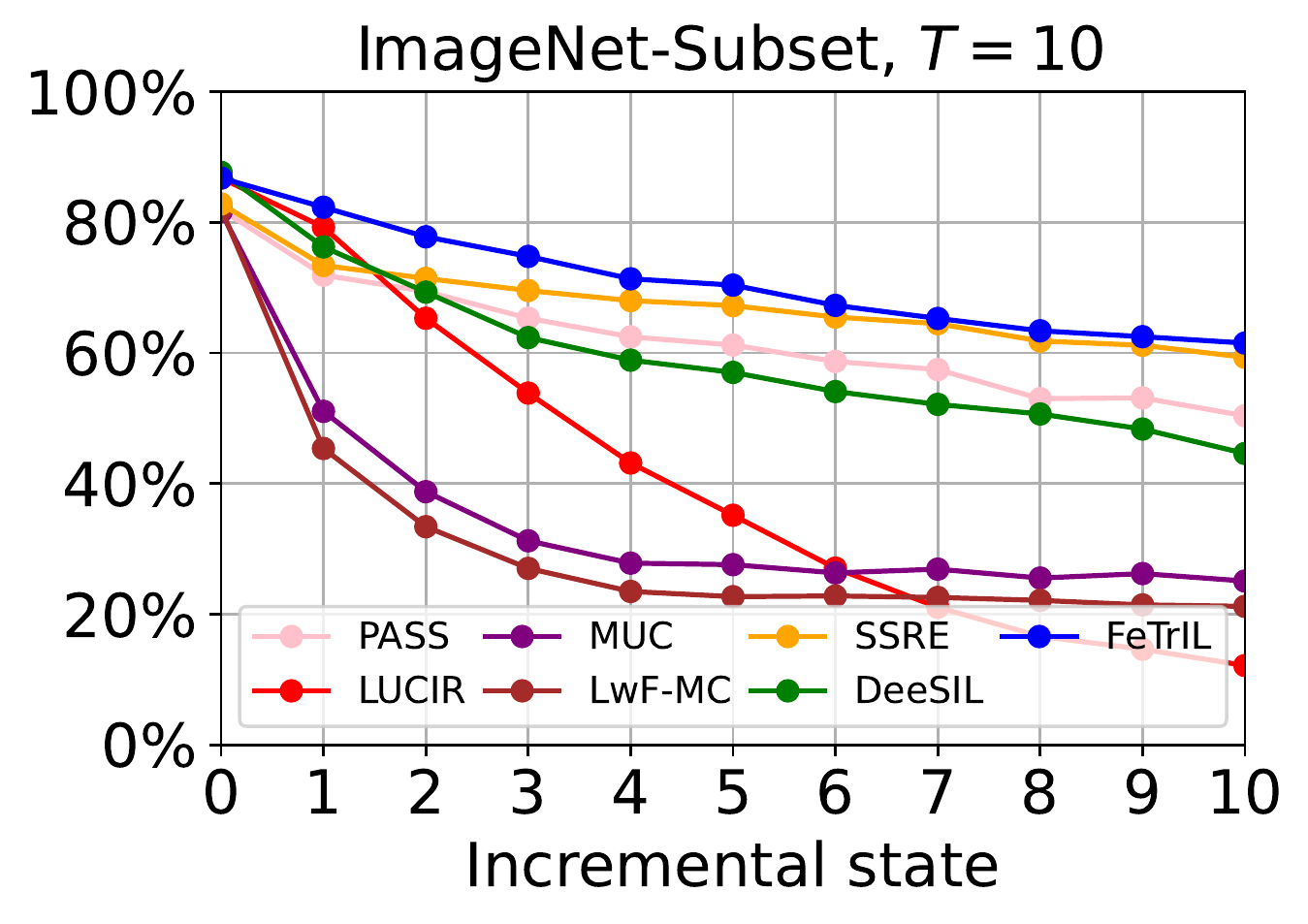} }}%
	\caption{Evolution of top-1 accuracy for an incremental process with $T=10$ IL states. \textit{Best viewed in color.} 
	}
\label{fig:accuracy}
\vspace{-2mm}
\end{figure*}

\textbf{Detailed view of accuracy.} We illustrate the evolution of accuracy across incremental states in Figure~\ref{fig:accuracy} to complement the averaged results from Table~\ref{tab:main_results}. 
These detailed results confirm the good behavior of the proposed method.
The evolution of accuracy for \ourmodel and SSRE is very similar for CIFAR-100, \ourmodel method is better throughout the process for TinyImageNet, and also better than SSRE for the first incremental states for ImageNet-Subset. 
The performance gain with respect to the other compared methods is much larger for all incremental states. 

\textbf{Comparison to exemplar-based CIL methods.} 
This comparison is interesting because EFCIL is a much more challenging task than EBCIL~\cite{belouadah2021_study,masana2021_study}, and an important performance gap between the two was observed.
This is intuitive since the storage of images of past classes in EBCIL mitigates catastrophic forgetting.
Following~\cite{hou2019_lucir,liu2021aanets}, a memory of 20 images per class is allowed for all EBCIL methods tested here.
\ourmodel is better than all three base methods to which AANets is applied for CIFAR-100. 
For ImageNet-Subset, \ourmodel accuracy is better than LUCIR's, slightly behind that of Mnemonics~\cite{mnemonics_2020} and approximately 3.5 points lower than that of PODNet~\cite{douillard2020podnet}.
The performance of \ourmodel remains close that of EBCIL methods in a majority of cases even after the introduction of AANets. 
The results from Table~\ref{tab:cil-mem} indicate that, while still present, the gap between EFCIL and EBCIL methods is narrowing. 

\begin{table}[t]
\begin{center}
\resizebox{0.47\textwidth}{!}{
\begin{tabular}{@{\kern0.5em}llccccccccccccl}
        \toprule
        \multirow{2}{*}{\textbf{CIL Method}}
            & \multicolumn{2}{c}{CIFAR-100}
            && \multicolumn{2}{c}{ImageNet-Subset}
            
        \\ \cmidrule(lr){2-3} \cmidrule(lr){5-6}  
        
            & \multicolumn{1}{c}{\textit{T}\ =\ 5}
            & \multicolumn{1}{c}{\textit{T}\ =\ 10}
            &
            & \multicolumn{1}{c}{\textit{T}\ =\ 5}
            & \multicolumn{1}{c}{\textit{T}\ =\ 10}
            
        \\ \midrule

LUCIR~\cite{hou2019_lucir} \small (CVPR'19) & 63.2 & 61.1  && 70.8 & 68.3  \\ 
+AAnets \small (CVPR'21) & 66.7 & 65.3  && 72.6 & 69.2  \\ 
\hline
Mnemonics~\cite{liu2020mnemonics} \small (CVPR'20) & 63.3 & 62.3  && 72.6 & 71.4  \\ 
+AAnets \small (CVPR'21) & 67.6 & 65.7  && 72.9 & 71.9  \\ 
\hline
PODNet~\cite{douillard2020podnet} \small (ECCV'20) & 64.8 & 63.2  && 75.5 & 74.3  \\ 
+AAnets \small (CVPR'21) & 66.3 & 64.3  && 77.0 & 75.6  \\ 
\hline
\ourmodelone &  66.3 & 65.2  && 71.9 & 70.8  \\     
\bottomrule
\end{tabular}
}
\end{center}
\vspace{-2mm}
	\caption{Comparison of \ourmodel with the recent AANets method~\cite{liu2021aanets}, applied on top of EBCIL baselines which store 20 exemplars of past classes to mitigate catastrophic forgetting. }
\label{tab:cil-mem}
\vspace{-4mm}
\end{table}

\subsection{Method analysis}
\label{subsec:analysis}
\vspace{-1mm}

We present an analysis of: (1) the selection strategies, (2) the memory footprint of the methods, (3) the complexity of model updates, and (4) the stability-plasticity balance. 

\begin{table}[t]
\begin{center}
\resizebox{0.47\textwidth}{!}{
\begin{tabular}{@{\kern0.5em}lcccccc@{\kern0.5em}}
     \toprule
       & CIFAR-100   & TinyImageNet & ImageNet-Subset \\
       \cmidrule(lr){2-4}
       & \multicolumn{3}{c}{\textit{T}\ =\ 5} \\
     \midrule

\ourmodelNospace$^{1}$       & 66.3 & 54.8 & 72.2\\
\ourmodelNospace$^{5}$       & 65.7 & 53.8 & 72.2\\
\ourmodelNospace$^{10}$      & 65.1 & 53.8 & 71.6\\
\ourmodelNospace$^{herd}$    & 66.2 & 53.8 & 72.1 \\
\ourmodelNospace$^{rand}$ & 65.1 & 51.5 & 70.3 \\

\midrule
\end{tabular}
}
\end{center}
\vspace{-2mm}
	\caption{Average top-1 CIL accuracy obtained with the variants of pseudo-feature selection from Subsection~\ref{subsec:selection} for $T=5$. We set $k=\{1,5,10\}$ for the similarity rank between the past and new classes to test the effect of class similarities. There are 10 (CIFAR-100 and ImageNet-Subset) and 20 (TinyImageNet) new classes per state from which to select features translation.  
	}
\vspace{-4mm}
\label{tab:selection}
\end{table}

\textbf{Pseudo-feature selection comparison.}
\ourmodel can use any past-new classes combination for translation.
In Table~\ref{tab:selection}, we compare the selection strategies from Subsection~\ref{subsec:selection}.
Accuracy varies in a relatively small range for all strategies, indicating that \ourmodel is robust to the way features of new classes are selected, and it can be successfully implemented with any of the strategies. 
\ourmodelNospace$^{1}$ is better than the other selection methods and this motivates its use in the main experiments.
Class similarity matters, but results with \ourmodelNospace$^{10}$ remain interesting.
\ourmodelNospace$^{herd}$ also has interesting accuracy, but is slightly behind that of \ourmodelNospace$^{1}$.
The results from Table~\ref{tab:selection} motivate the use of \ourmodelNospace$^{1}$ in the main experiments.
Overall, the geometric translation toward the centroid of the past class is by far more important than the new classes features sampling policy.
This finding is also supported by the results obtained with a single new class per CIL state (Table~\ref{tab:main_results}).

\textbf{Memory footprint.} 
A low memory footprint is a desirable property of incremental learning algorithms because they are most useful in memory-constrained applications~\cite{masana2021_study,ravaglia2021tinyml,rebuffi2017_icarl}, and recommended for embedded devices~\cite{hayes2022online}.
All EFCIL methods need to store a representation of past classes to counter catastrophic forgetting. 
Naturally, this representation should be as compact as possible. 
Mainstream methods (such as LwF-MC~\cite{li2016_lwf}, PASS~\cite{zhu2021pass}, IL2A~\cite{zhu2021pass}, and SSRE~\cite{zhu2022self}) need to the previous and current deep models during CIL updates for distillation.
ResNet-18~\cite{he2016_resnet}, the most frequent CIL backbone, has approximately 11.4M parameters. 
Consequently, distillation-based methods require around 22.8M parameters.
Transfer-based methods, such as DeeSIL~\cite{belouadah2018_deesil} and \ourmodel, use only the deep model learned in the initial state and frozen afterwards, and only need 11.4M parameters for the model.
DeeSIL does not need supplementary parameters during incremental updates.
However, this comes at the cost of poor global discrimination of classes, which is reflected in the final performance. 
\ourmodel stores the class centroids of past classes in order to perform feature translation.
Each class needs 512 parameters, which leads to a supplementary 51.2K and 102.4 memory need for 100 and 200 classes, respectively.
The class similarities needed for pseudo-feature selection (Subsection~\ref{subsec:selection}) can be computed sequentially and the added memory cost of this step is negligible.
PASS~\cite{zhu2021pass}, IL2A~\cite{zhu2021class} and SSRE~\cite{zhu2022self} also require the storage of a prototype (mean representation) for each past class and their footprint is equivalent to that of \ourmodel. 
IL2A~\cite{zhu2021class} additionally stores a covariance matrix per past class (512x512 for ResNet-18) for optimal functioning, which is prohibitive. 

\begin{figure}
\centering
    {\includegraphics[width=.6\linewidth]{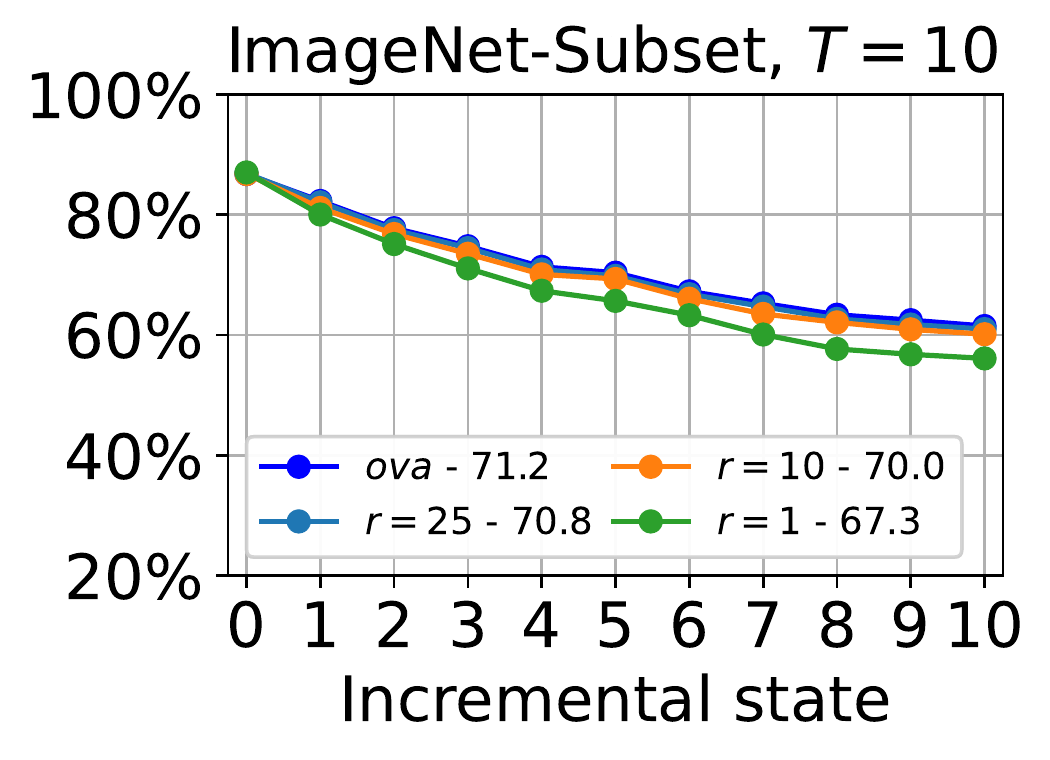} }
\vspace{-1mm}
\caption{Top-1 incremental accuracy of \ourmodelone for approximate training of the classification layer with different ratios for negative sampling. $ova$ denotes a classical one-vs-all training procedure which is used to report the main results from Table~\ref{tab:main_results} and Figure~\ref{fig:accuracy}. 
}
\vspace{-5mm}
\label{fig:ratios}
\end{figure}

\textbf{Complexity of incremental updates.}
CIL is useful in resource-constrained environments, and the integration of new classes should be fast~\cite{hayes2022online,ravaglia2021tinyml}.
Distillation-based methods retrain the full backbone model at each update.
This is costly because backpropagation complexity depends on the network architecture, the number of samples and the number of epochs~\cite{goodfellow2016deep}.
Updates of transfer-based methods are simpler because they update only the final layer.
DeeSIL trains linear classifiers using a one-vs-all procedure within each CIL state.
The complexity of one training epoch for all classifiers in a CIL state is $O((\frac{n}{T})^2sd)$~\cite{bottou2007tradeoffs}, with $n$ - total number of classes in the dataset, $d$ - dimensionality of features and $s$ - samples per class.
\ourmodel retrains all linear classifiers, past and new, in each CIL state to improve global separability. 
Its complexity is $O(n^2sd)$ in the last incremental state, which includes all classes.
However, the one-versus-all training can be replaced with a one-versus-many training with negligible loss of accuracy.
A sampling of negative features is performed to respect a predefined ratio $r$ between negatives and positives used to train each classifier.
This approximation has $O(rnsd)$ complexity. 
It is interesting since $r<n$, and is more and more useful as $n$ grows during the IL process since $r$ remains constant.

In Figure~\ref{fig:ratios}, we present results with different $r$ values for ImageNet-Subset, $T=10$. 
Accuracy drops when negative sampling is performed, but it is close to that of one-vs-all training when $r=25$ and $r=10$.
Performance drops more significantly for $r=1$, when each linear classifier is learned with an aggressive sampling of negatives.
Similar results for CIFAR-100 and TinyImageNet are provided in the suppl. material.
Globally, Figure~\ref{fig:ratios} indicates that \ourmodel increments can be accelerated with little accuracy loss. 

We measure the time needed for incremental training of ImageNet-Subset, $T=10$.
The training of the initial model is similar for all models and is thus discarded.
\ourmodel training is done on a single thread of an Intel E5-2620v4 CPU, and only takes 1 hour, 4 minutes and 16 seconds. 
If \ourmodel is run with $r=10$ ratio between positives and negatives, training time is only 15 minutes and 3 seconds. 
In comparison, PASS~\cite{zhu2021pass} needs 11 hours, 8 minutes and 19 seconds on an NVIDIA V100 GPU, with 4 workers for data loading.
While clearly favorable to \ourmodelNospace, the comparison is biased in favor of PASS since this method uses an entire GPU, in comparison to a single CPU thread for \ourmodel. 
Further speed gains are possible for our method by using a GPU implementation of the linear layer.
Our method would run much faster with a GPU implementation of the linear layer. 
Note that the running time of the other methods, such as LUCIR~\cite{hou2019_lucir} and SSRE~\cite{zhu2022self},  which perform backpropagation is similar to that of PASS~\cite{zhu2021pass}. 

\begin{figure}
\centering
    \includegraphics[width=.996\linewidth]{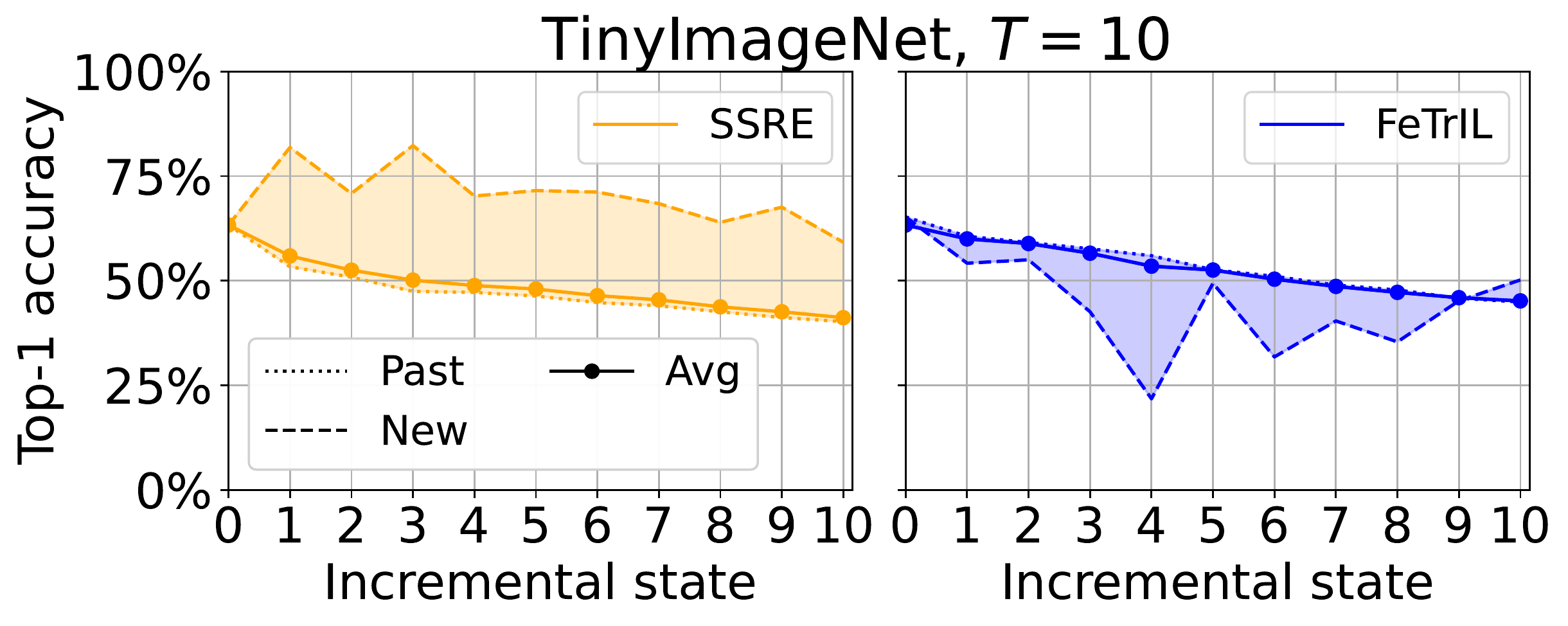}\\
\caption{Top-1 incremental accuracy per state for past and new classes for TinyImageNet, with $T=10$ incremental states for \ourmodelone and SSRE, the best compared method. An ideal method would provide high accuracy, but also similar performance for past and new classes. The accuracy of past and new classes is globally closer for \ourmodelone, which indicates that our method provides a better stability-plasticity balance than SSRE. Overall accuracy is better for \ourmodelone in Figure~\ref{fig:accuracy} because the contribution of new classes in each state diminishes during the CIL process.
}
\vspace{-4mm}
\label{fig:balance}
\end{figure}

\textbf{Stability-plasticity balance.}
CIL should ideally ensure a similar accuracy level for past and new classes~\cite{masana2021_study,zhu2022self}.
Figure~\ref{fig:balance} shows that the two methods have complementary behavior, which results from the way deep backbones are used.
SSRE is biased toward new classes since the model is fine tuned in each incremental state.
\ourmodel favors past classes because the deep model is learned with the initial classes (a subset of past classes) and then frozen.
The accuracy gap between past and new classes is smaller for \ourmodel compared to SSRE, except for state 4.
There, low performance on new classes is probably explained by a strong domain shift compared to the initial state. 
Globally, the proposed method improves the stability-plasticity balance.

\section{Conclusion}
\label{sec:conclusions}
We introduce \ourmodel, a new method which addresses exemplar-free class-incremental learning. 
The proposed combination of a frozen feature extractor and of a pseudo-feature generator improves results compared to recent EFCIL methods.
The generation of pseudo-features is simple, since it consists in a geometric translation, yet effective.
Our proposal is advantageous from memory and speed perspectives compared to mainstream methods~\cite{hou2019_lucir,rebuffi2017_icarl,smith2021always,kumar2021_efficient,sdc_2020,zhu2021class,zhu2021pass,zhu2022self}.
This is particularly important for edge devices~\cite{hayes2022online,ravaglia2021tinyml}, whose storage and computation capacities are limited.
\ourmodel performance is also close to that of exemplar-based methods, which need to store samples of past classes to mitigate catastrophic forgetting.
While a gap between exemplar-based and exemplar-free setting subsists, it becomes significantly narrower. 
The results reported here resonate with past works which show that simple methods can be highly effective in CIL~\cite{belouadah2021_study,masana2021_study,prabhu2020gdumb}.
They question the usefulness of the knowledge distillation component, used by a majority of existing methods.
The \ourmodel code will be made public to enable reproducibility. 

The main limitations of the proposed method motivate our future work.
First, \ourmodel uses a frozen feature extractor learned on the initial state and tends to favor past classes over new ones. 
We will investigate ways to combine the pseudo-feature generation mechanism and fine-tuning to further improve global performance, as well as the stability-plasticity balance.
Second, \ourmodel produces usable pseudo-features, but past class representations would be better if the pseudo-features would be more similar to the original features of past classes. 
We will study methods that generate more refined features, for instance by using the distribution of the initial features.
Last but not least, the tested selection strategies are all effective.
However, they could be further improved by filtering out outliers based on the localization of pseudo-features in the representation space. 

\noindent \textbf{Acknowledgements}.
This work was supported by the European Commission under European Horizon 2020 Programme, grant number 951911 - AI4Media. It was made possible by the use of the FactoryIA supercomputer, financially supported by the Ile-de-France Regional Council.

{\small
\bibliographystyle{ieee_fullname}
\bibliography{egbib}
}

\end{document}